\documentclass{article}

\usepackage[final]{corl_2017} 

\usepackage[utf8]{inputenc} 
\usepackage[T1]{fontenc}    
\usepackage{url}            
\usepackage{booktabs}       
\usepackage{amsfonts}       
\usepackage{hyperref}
\usepackage{amsmath}        
\usepackage{nicefrac}       
\usepackage{graphicx}       
\usepackage{microtype}      
\usepackage{amsmath}
\usepackage[font={footnotesize}]{caption}
\usepackage{enumitem}

\usepackage{rotating}
\usepackage{array}

\usepackage{floatrow}
\usepackage{subcaption}
\newfloatcommand{capbtabbox}{table}[][\FBwidth]

\setlist{nosep}

\usepackage{color}

\title{Sim-to-Real Robot Learning from Pixels with Progressive Nets}

\author{
Andrei A. Rusu\\
DeepMind \\
London, UK\\
\texttt{andreirusu@google.com} \\
\And
Mel Večerík\\
DeepMind \\
London, UK\\
\texttt{matejvecerik@google.com} \\
\And
Thomas Rothörl\\
DeepMind \\
London, UK\\
\texttt{tcr@google.com} \\
\And
Nicolas Heess\\
DeepMind \\
London, UK\\
\texttt{heess@google.com} \\
\And
Razvan Pascanu\\
DeepMind \\
London, UK\\
\texttt{razp@google.com} \\
\And
Raia Hadsell \\
DeepMind \\
London, UK\\
\texttt{raia@google.com} \\
}

\begin{document}
\maketitle


\begin{abstract}
Applying end-to-end learning to solve complex, interactive, pixel-driven control tasks on a robot is an unsolved problem. Deep Reinforcement Learning algorithms are too slow to achieve performance on a real robot, but their potential has been demonstrated in simulated environments. We propose using \emph{progressive networks} to bridge the reality gap and transfer learned policies from simulation to the real world. The progressive net approach is a general framework that enables reuse of everything from low-level visual features to high-level policies for transfer to new tasks, enabling a compositional, yet simple, approach to building complex skills. We present an early demonstration of this approach with a number of experiments in the domain of robot manipulation that focus on bridging the reality gap. Unlike other proposed approaches, our real-world experiments demonstrate successful task learning from raw visual input on a fully actuated robot manipulator. Moreover, rather than relying on model-based trajectory optimisation, the task learning is accomplished using only deep reinforcement learning and sparse rewards. 
\end{abstract}



\section{Introduction}

Deep Reinforcement Learning offers new promise for achieving human-level control in robotics domains, especially for pixel-to-action scenarios where state estimation is from high dimensional sensors and environment interaction and feedback are critical. With deep RL, a new set of algorithms has emerged that can attain sophisticated, precise control on challenging tasks, but these accomplishments have been demonstrated primarily in simulation, rather than on actual robot platforms. 

While recent advances in simulation-driven deep RL are impressive \cite{Levine2014Guided,Schulmanetal_ICML2015,Heess2015Stochastic,Lillicrap2016Continuous,Schulmanetal_ICLR2016,mnih2016a3c,Gu2016NAF}, demonstrating learning capabilities on real robots remains the bar by which we must measure the practical applicability of these methods. However, this poses a significant challenge, given the "data-hungry" training regime required for current pixel-based deep RL methods, and the relative frailty of research robots and their human handlers. One solution is to use transfer learning methods to bridge the reality gap that separates simulation from real world domains. In this paper, we use progressive networks, a deep learning architecture that has recently been proposed for transfer learning, to demonstrate such an approach, thus providing a proof-of-concept pathway by which deep RL can be used to effect fast policy learning on a real robot.

Progressive nets have been shown to produce positive transfer between disparate tasks such as Atari games by utilizing lateral connections to previously learnt models \cite{rusu2016progressive}. The addition of new capacity for each new task allows specialized input features to be learned, an important advantage for deep RL algorithms which are improved by sharply-tuned perceptual features. An advantage of progressive nets compared with other methods for transfer learning or domain adaptation is that multiple tasks may be learned sequentially, without needing to specify source and target tasks.

This paper presents an approach for transfer from simulation to the real robot that is proven using real-world, sparse-reward tasks. The tasks are learned using end-to-end deep RL, with RGB inputs and joint velocity output actions. First, an actor-critic network is trained in simulation using multiple asynchronous workers \cite{mnih2016a3c}. The network has a convolutional encoder followed by an LSTM. From the LSTM state, using a linear layer, we compute  a set of discrete action outputs that control the different degrees of freedom of the simulated robot as well as the value function. After training, a new network is initialized with lateral, nonlinear connections to each convolutional and recurrent layer of the simulation-trained network. The new network is trained on a similar task on the real robot. Our initial findings show that the inductive bias imparted by the features and encoded policy of the simulation net is enough to give a dramatic learning speed-up on the real robot.

\section{Transfer Learning from Simulation to Real}

Our approach relies on the \emph{progressive nets} architecture, which enables transfer learning through lateral connections which connect each layer of previously learnt network columns to each new column, thus supporting rich compositionality of features. We first summarize progressive nets, and then we discuss their application for transfer in robot domains. 

\subsection{Progressive Networks}

Progressive networks are ideal for simulation-to-real transfer of policies in robot control domains, for multiple reasons. First, features learnt for one task may be transferred to many new tasks without destruction from fine-tuning. Second, the columns may be heterogeneous, which may be important for solving different tasks, including different input modalities, or simply to improve learning speed when transferring to the real robot. Third, progressive nets add new capacity, including new input connections, when transferring to new tasks. This is advantageous for bridging the reality gap, to accommodate dissimilar inputs between simulation and real sensors. 


A progressive network starts with a single column: a deep neural network having $L$ layers with hidden activations $h_i^{(1)} \in \mathbb{R}^{n_i}$, with $n_i$ the number of units at layer $i \le L$, and parameters $\Theta^{(1)}$ trained to convergence.  When switching to a second task, the
parameters $\Theta^{(1)}$ are ``frozen'' and a new
column with parameters $\Theta^{(2)}$ is instantiated (with random initialization), where layer $h_i^{(2)}$ receives input from both $h_{i-1}^{(2)}$ and $h_{i-1}^{(1)}$ via lateral connections. Progressive networks can be generalized in a straightforward manner to have arbitrary network width per column/layer, to accommodate varying degrees of task difficulty, or to compile lateral connections from multiple, independent networks in an ensemble setting. 
\begin{align}
  \label{eq:prognet}
  h_i^{(k)} = f\left( W_i^{(k)} h_{i-1}^{(k)} + \sum_{j<k} U_{i}^{(k:j)} h_{i-1}^{(j)} \right),
\end{align}
where $W_i^{(k)} \in \mathbb{R}^{n_{i} \times n_{i-1}}$ is the weight matrix of layer
$i$ of column $k$, $U_{i}^{(k:j)} \in \mathbb{R}^{n_i \times n_j}$ are the
lateral connections from layer $i-1$ of column $j$, to layer $i$ of column $k$
and $h_0$ is the network input.
$f$ is an element-wise non-linearity: we use $f(x)=\max(0, x)$ for all
intermediate layers.

In the standard pretrain-and-finetune paradigm, there is often an implicit assumption of ``overlap'' between the tasks. Finetuning is efficient in this setting, as parameters need only be adjusted slightly to the target domain, and often only the top layer is retrained. In contrast, we make no assumptions about the relationship between tasks, which may in practice be orthogonal or even adversarial. Progressive networks side-step this issue by allocating a new column, potentially with different structure or inputs, for each new task. Columns in progressive networks are free to reuse, modify or ignore previously learned features via the lateral connections.

\paragraph{Application to Reinforcement Learning.} Although progressive networks are widely applicable, this paper focuses on their application to deep reinforcement learning. In this case, each column is trained to solve a particular Markov Decision Process (MDP): the $k$-th column thus defines a policy $\pi^{(k)}(a\mid s)$ taking as input a state $s$ given by the environment, and generating probabilities over actions $\pi^{(k)}(a\mid s) := h_L^{(k)}(s)$.
At each time-step, an action is sampled from this distribution and taken in the environment, yielding the subsequent state. This policy implicitly defines a stationary distribution $\rho_{\pi^{(k)}}(s,a)$ over states and actions.

\subsection{Approach}

\begin{figure}[h]
  \centering
    \includegraphics[width=.6\textwidth]{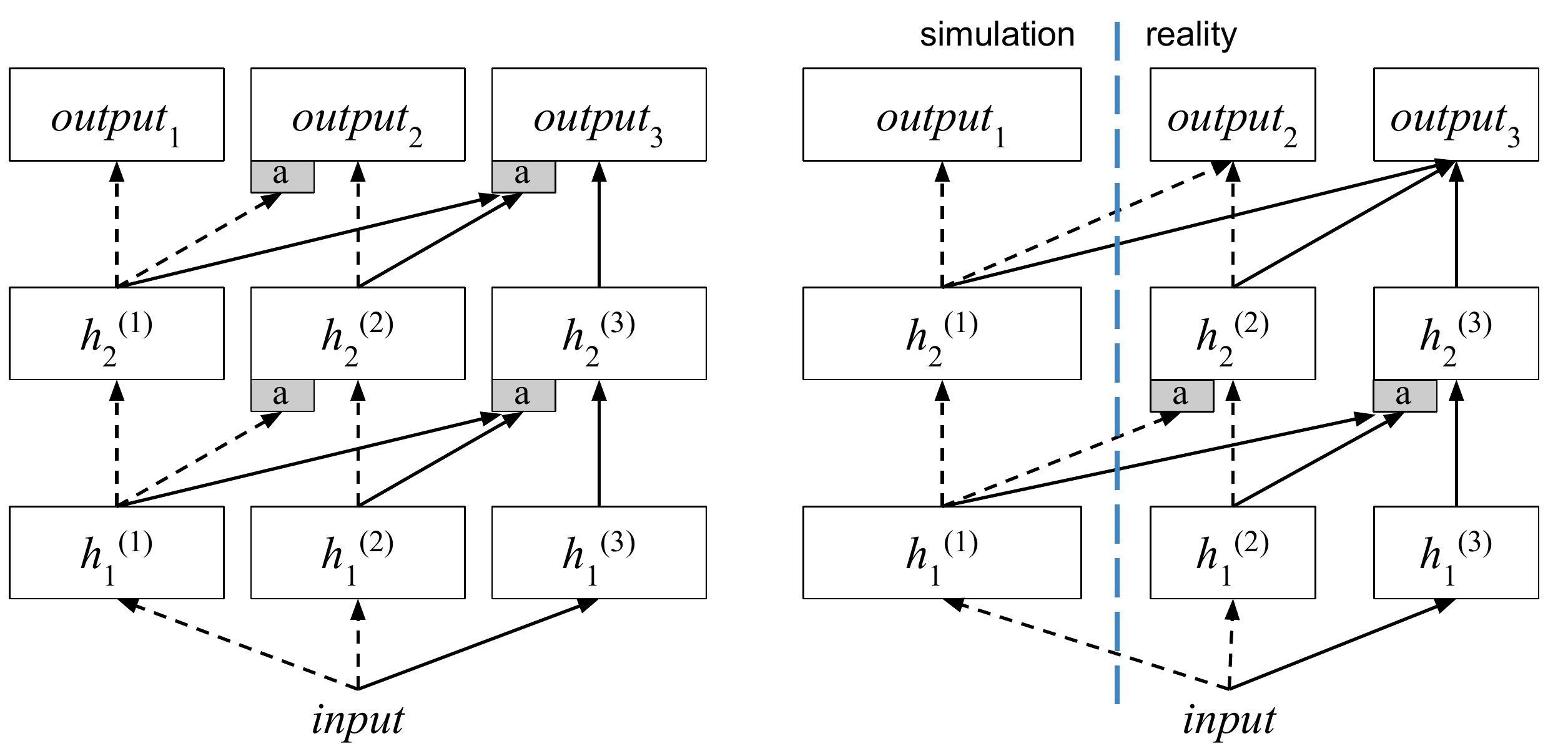}
    \caption{Depiction of a progressive network, left, and a modified progressive architecture used for robot transfer learning, right. The first column is trained on Task 1, in simulation, the second column is trained on Task 1 on the robot, and the third column is trained on Task 2 on the robot. Columns may differ in capacity, and the adapter functions (marked `a') are not used for the output layers of this non-adversarial sequence of tasks.
    }
    \label{fig:progressiveSim2Real}
\end{figure}

The proposed approach for transfer from simulated to real robot domains is based on a progressive network with some specific changes. First, the columns of a progressive net do not need to have identical capacity or structure, and this can be an advantage in sim-to-real situations. Thus, the simulation-trained column is designed to have sufficient capacity and depth to learn the task from scratch, but the robot-trained columns have minimal capacity, to encourage fast learning and limit total parameter growth. Secondly, the layer-wise adapters proposed for progressive nets are unnecessary for the output layers of complementary sequences of tasks, so they are not used. Third, the output layer of the robot-trained column is initialised from the simulation-trained column in order to improve  exploration. These architectural features are shown in Fig. \ref{fig:progressiveSim2Real}.

The greatest risk in this approach to transfer learning is that rewards will be so sparse, or non-existent, in the real domain that the reinforcement learning will not improve a vastly suboptimal initial policy within a practical time frame. Thus, in order to maximise the likelihood of reward during exploration in the real domain, the new column is initialised such that the initial policy of the agent will be identical to the previous column. This is accomplished by initialising the weights coming from the last layer of the previous column to the  output layer of the new column with the output weights of the previous column, and the connections incoming from the  last hidden layer of the current column are initialised with zero-valued weights. Thus, using the example network in Fig. \ref{fig:progressiveSim2Real} (right), when parameters $\Theta^{(2)}$ are instantiated, layer $output_2^{(2)}$ will have input connections from $h_2^{(1)}$ and $h_2^{(2)}$. However, unlike the other parameters in  $\Theta^{(2)}$, which will be randomly initialised, the weights $W_{out}^{(2)}$ will be zeros and the weights $U_{out}^{(1:2)}$ will be copied from $W_{out}^{(1)}$. Note that this only affects the initial policy of the agent and does not prevent the new column from training. 

\section{Related Literature}

There exist many different paradigms for domain transfer and many approaches designed specifically for deep neural models, but
substantially fewer approaches for transfer from simulation to reality for robot domains. Even more rare are methods that can be used for transfer in interactive, rich sensor domains using end-to-end (pixel-to-action) learning.

A growing body of work has been investigating the ability of deep networks to transfer between domains. Some research \cite{PengSAS15, SuQLG15} considers simply augmenting the target domain data with data from the source domain where an alignment exists. Building on this work, \cite{LongC0J15} starts from the 
observation that as one looks at higher layers in the model, the transferability of the features decreases quickly. To correct this effect, 
 a soft constraint is added that enforces the distribution 
of the features to be more similar. 
In \cite{LongC0J15}, a `confusion' loss is proposed which forces 
the model to ignore variations in the data that separate the two domains \cite{TzengHDS15, TzengHZSD14}. 

Based on \cite{TzengHDS15}, \cite{TzengDHFPLSD15} attempts to address the simulation to reality gap by using aligned data. The work is focused on pose estimation of the robotic arm, where training happens on a triple loss that looks at aligned simulation to real data, including the domain confusion loss. The paper does not show the efficiency of the method on learning novel complex policies.

Several recent works from the supervised learning literature, e.g.\ \cite{ganin2015domain,ajakanDomain2016,BousmalisTSKE16}, demonstrate how ideas from the adversarial training of neural networks can be used to reduce the sensitivity of a trained network to inter-domain variations, without requiring aligned training data. Intuitively these approaches train a representation that makes it hard to distinguish between data points drawn from the different domains. These ideas have, however, not yet been tested in the context of control.
Demonstrating the difficulty of the problem, \cite{SuQLG15} provides evidence that a simple application of a model trained on synthetic data on the real robot fails. The paper also shows that the main failure point is the discrepancy in visual cues between simulation and reality.

Partial success on transferring from simulation to a real robot has been reported \cite{AAMASWS10-barrett, 2016JamesJohns, ZhuMKLGFF16}. They focus primarily on the problem of transfer from a more restricted simpler version of a task to the full, more difficult version. 
While transfer from simulation to reality remains difficult, progress has been made with directly learning neural network control policies on a real robot, both from low-dimensional representations of the state and from visual input (e.g. \cite{LevineFDA15},\cite{LevineWA15}). While the results are impressive, to achieve sufficient data efficiency these works currently rely on relatively restrictive task setups, specialized visual architectures, and carefully designed training regimes. Alternative approaches embrace big data ideas for robotics (\cite{PintoG16,levine2016Large}).

\section{Experiments}

For training in simulation, we use the Asynchronous Advantage Actor-Critic (A3C) framework introduced in
\citep{mnih2016a3c}.  Compared to DQN \citep{mnih-dqn-2015}, the model simultaneously learns a policy and a value function
for predicting expected future rewards, and can be trained with CPUs, using multiple threads. A3C has been shown to converge faster than DQN, which makes it advantageous for research experimentation. 

For the manipulation domain of the Jaco arm, the agent policy controls nine degrees of freedom using velocity commands. This includes six joints on the arm plus three actuated fingers. The full policy $\Pi(\mathbf{A}|s,\theta)$ comprises nine joint policies learnt by the agent, each one a softmax connected to the inputs from the previous layer and any lateral connections. Each joint policy $i$ has three actions (a fixed positive velocity, a fixed negative velocity, and a zero velocity): $\pi_i(a_i|s;\theta_i)$. This discrete action set, while potentially lacking the precision of a continuous control policy, has worked well in practice. There is also a single value function that is linearly connected to the previous layer and lateral layers: $V(s,\theta_v)$. 

\begin{figure}
\begin{floatrow}
\ffigbox{%
    \includegraphics[width=.3\textwidth]{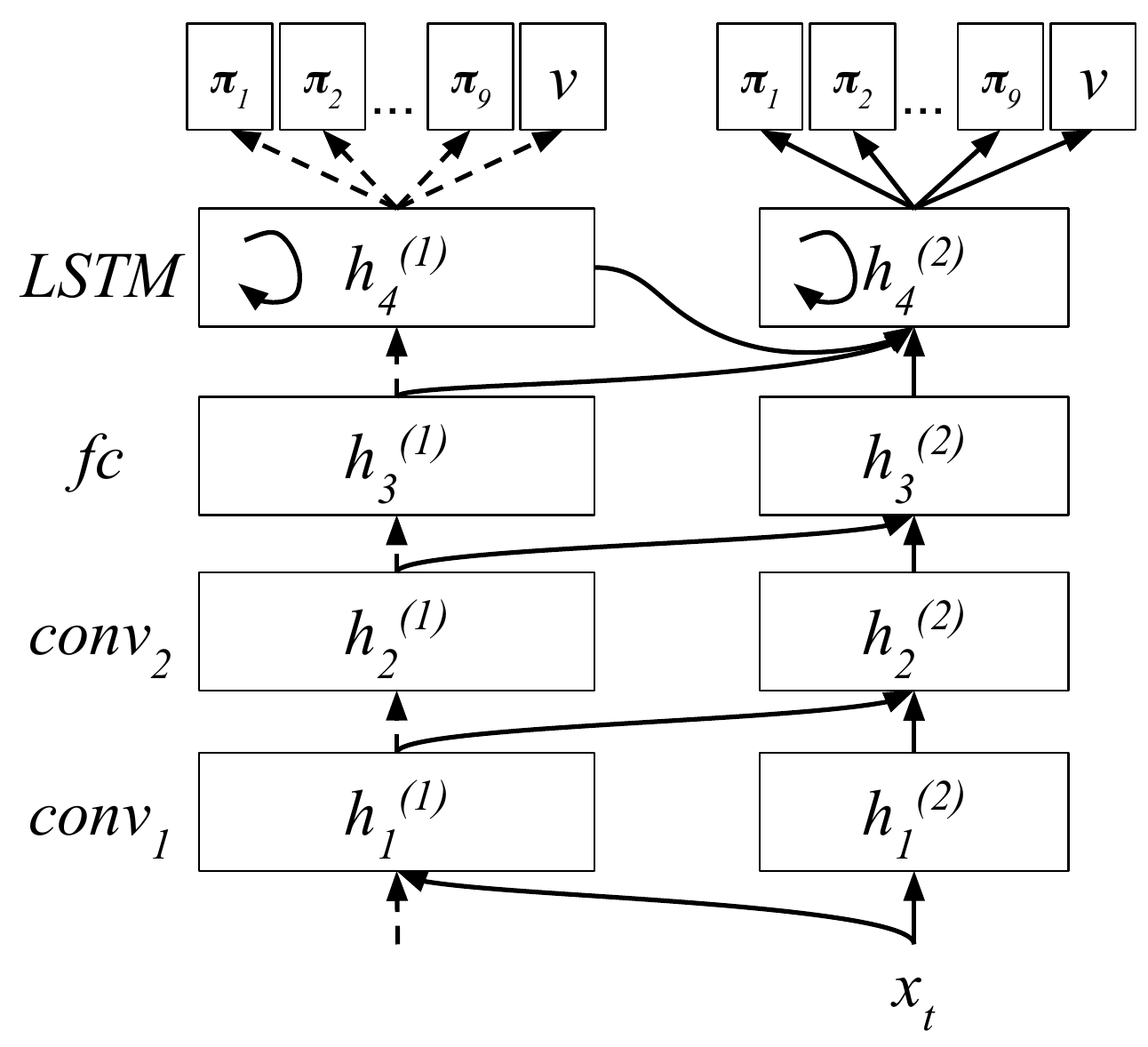}
}{%
    \caption{Detailed schematic of progressive recurrent network architecture. The activations of the LSTM are connected as inputs to the progressive column. The factored policy and single value function are shown.}
    \label{fig:networks}
}
\capbtabbox{%
\small
\begin{tabular}{@{}lrrrr@{}}
 \toprule
& \multicolumn{2}{c}{\textbf{feedforward}}
& \multicolumn{2}{c}{\textbf{recurrent}}\\
                    & wide    & narrow  & wide    & narrow  \\
 \midrule
  fc (output)    & 28    & 28  & 28   & 28    \\
  LSTM         & -     & -   & 128  & 16    \\
    fc       & 512   & 32  & 128  & 16    \\
  conv 2      & 32    & 8   & 32   & 8    \\
  conv 1      & 16    & 8   & 16   & 8   \\
  params & 621K  & 39K  & 299K & 37K \\
  
 \bottomrule
 \\
 \\
\end{tabular}
\caption{Network sizes for wide columns (simulation-trained) and narrow columns (robot-trained). For all networks, the first convolutional layer uses 8x8, stride 4 kernels and the second uses 5x5, stride 2 kernels. The total parameters include the lateral connections.}
    \label{table:network_specs}
}

\end{floatrow}
\end{figure}

We evaluate both feedforward and recurrent neural networks. Both have convolutional input layers followed by either a fully connected layer or an LSTM. A standard-sized network is used for the simulation-trained column and a reduced-capacity network is used for the robot-trained columns, chosen because we found empirically that more capacity does not accelerate learning (see Section\ref{sec:transfer}), presumably because of the features reused from the previous column. Details of the architecture are given in Figure \ref{fig:networks} and Table \ref{table:network_specs}. In all variants, the input is 3x64x64 pixels and the output is 28 (9 discrete joint policies plus one value function).

\begin{figure}
  \centering
    \includegraphics[width=.23\textwidth]{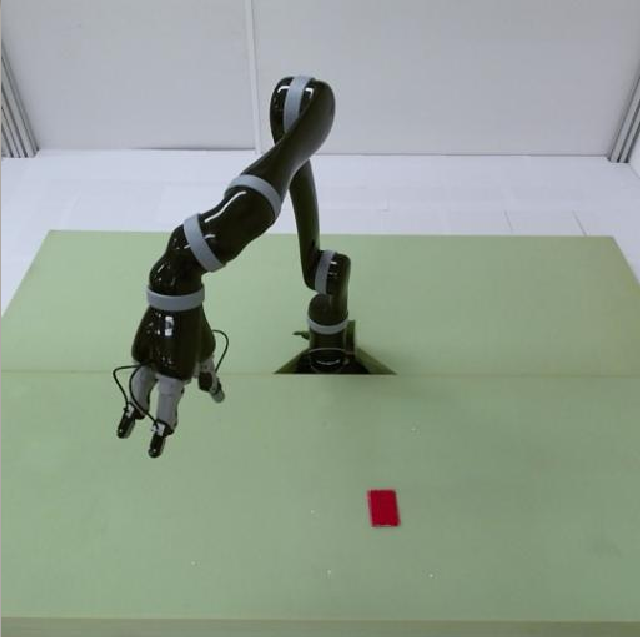}\hspace{.1in}
    \includegraphics[width=.23\textwidth]{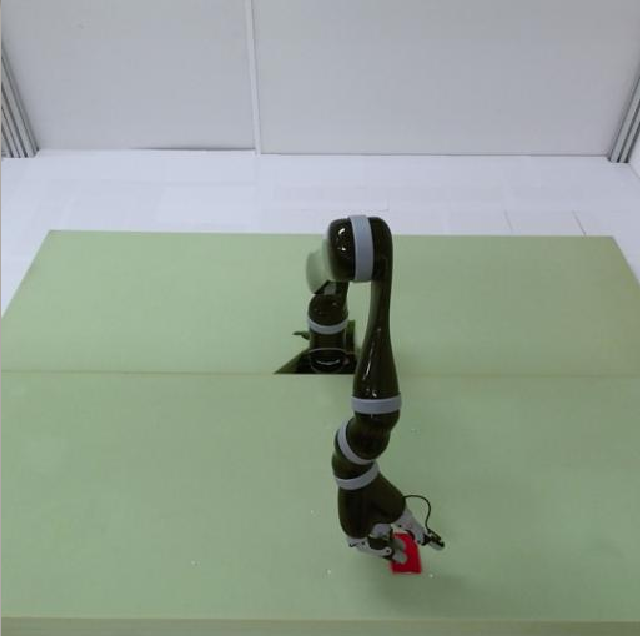}\hspace{.1in}
    \includegraphics[width=.23\textwidth]{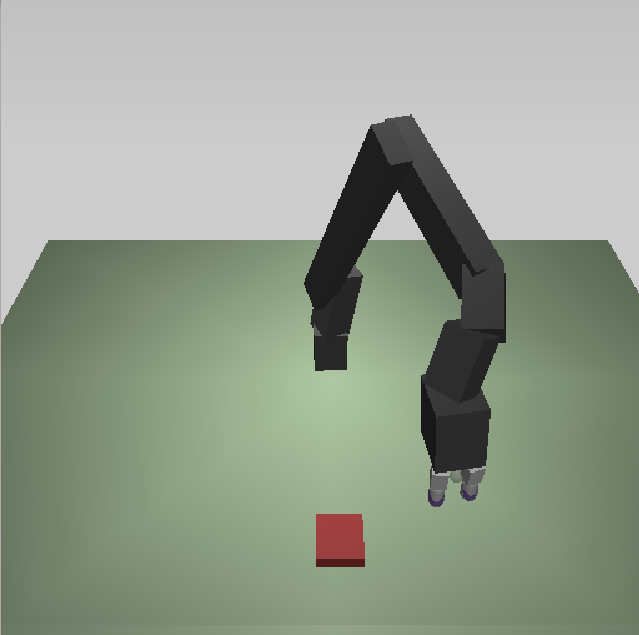}\hspace{.1in}
    \includegraphics[width=.23\textwidth]{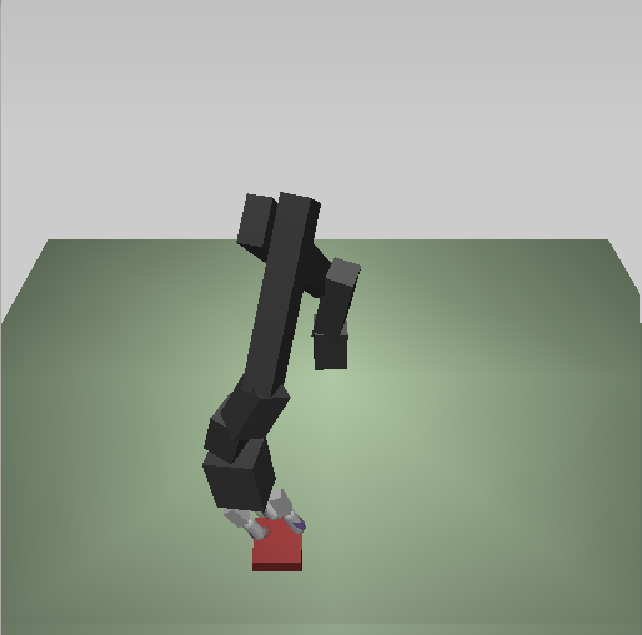}
    \caption{Sample images from the real camera input image and the MuJoCo-rendered image. Though a more realistic model appearance could have been used, the blocky Jaco model was used to accelerate MuJoCo rendering, which was done on CPUs. The images show the diversity of Jaco start positions and target positions.
    }
    \label{fig:jacoarm}
\end{figure}

The MuJoCo physics simulator \cite{Todorov2012} is used to train the first column for our experiments, with a rendered camera view to provide observations. In the real domain, a similarly positioned RGB camera provides the input. While the modeled Jaco and its dynamics are quite accurate, the visual discrepancies are obvious, as shown in Figure \ref{fig:jacoarm}. 

The experiments are all focused around the task of reaching to a visual target, with only pure rewards provided as feedback (no shaped rewards). Though simple, this task requires that the state of the arm and the position of the target are correctly inferred from visual observations, and that the agent learns robust control over a high-dimensional state space. The arm is set to a random start position at the beginning of every episode, and the target is placed randomly within a 40cm by 30cm area. The agent receives a reward of $+1$ if its palm is within 10cm of the target, and episodes last for at most 50 steps. Though there is some variance due to randomized starting states, a well-performing agent can achieve an average score of over 30 points by quickly reaching to the target and remaining in safe positions at all times. The episode is terminated if the agent causes a safety violation through self-intersection, by touching the table top, or by exceeding set joint limits.

\subsection{Training in simulation}

\begin{figure}
  \centering
    \includegraphics[width=.42\textwidth,clip]{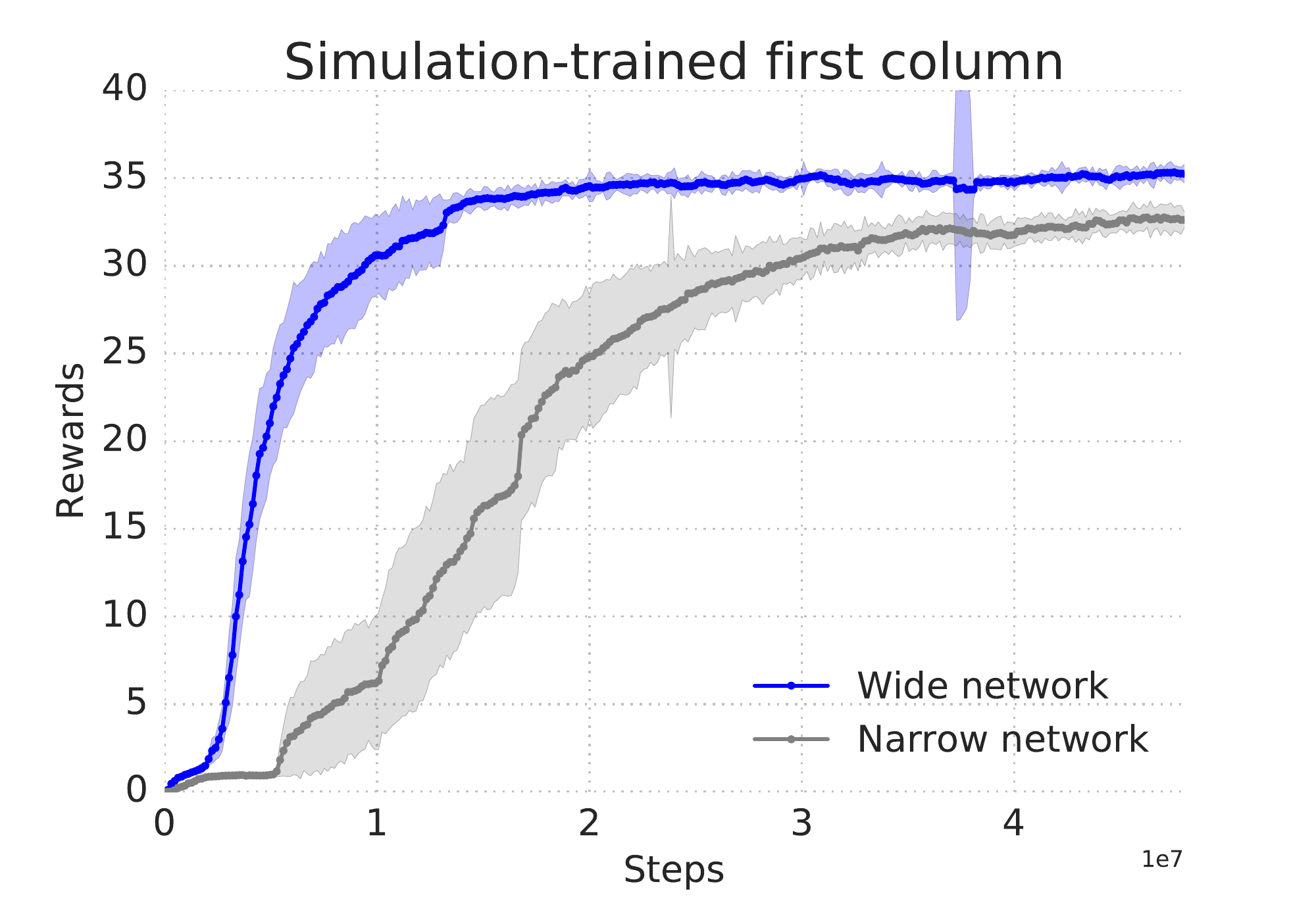}
    \hspace{.1in}
    \includegraphics[width=.42\textwidth,clip]{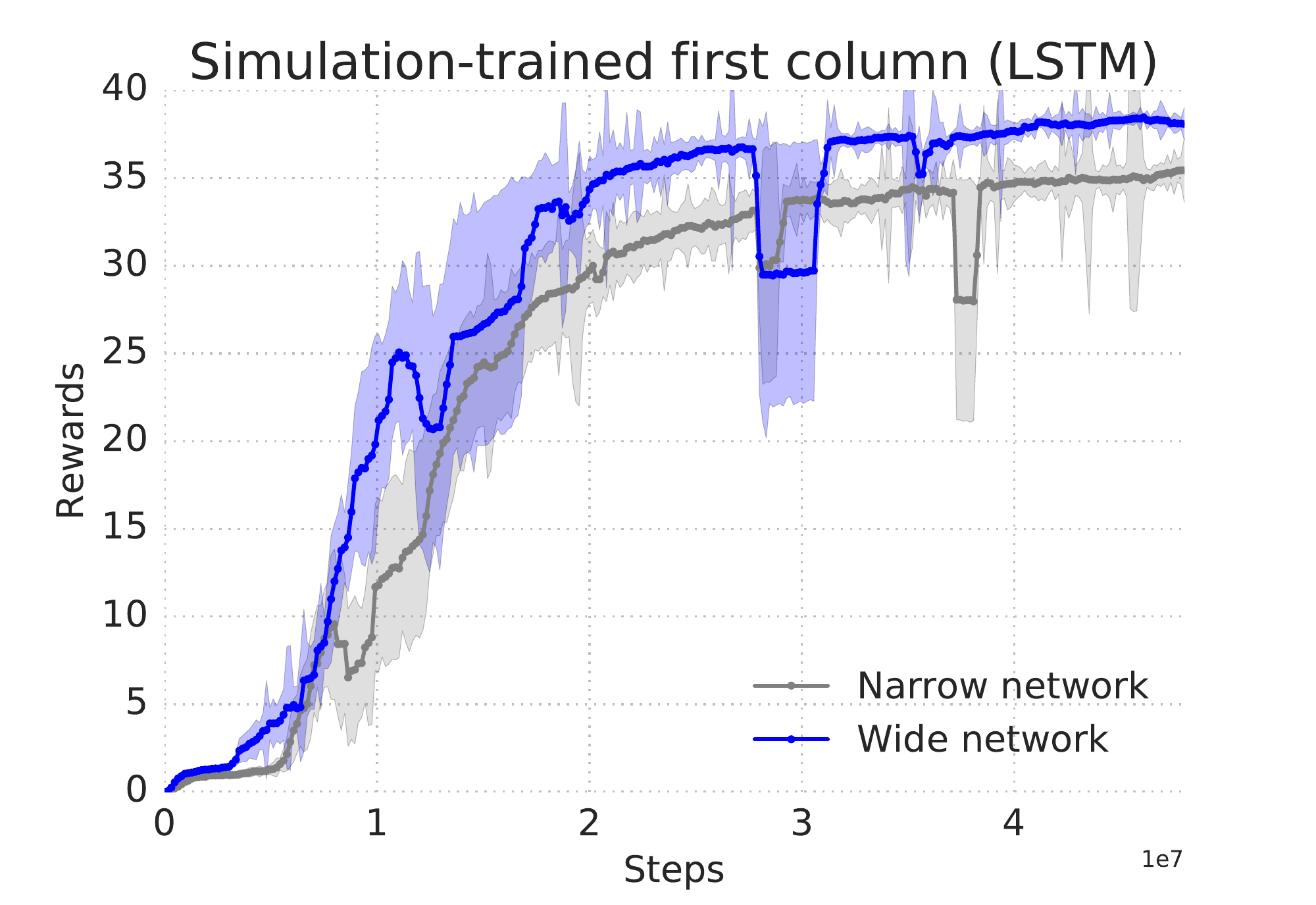}
    \caption{Learning curves are shown for wide and narrow versions of the feedforward (left) and recurrent (right) models, which are trained with the MuJoCo simulator. The plots show mean and variance over 5 training runs with different seeds and hyperparameters. Stable performance is reached after approximately 50 million steps, which is more than one million episodes. While both the feedforward and the recurrent models learn the task, the recurrent network reaches a higher final mean score.}
    \label{fig:widevsnarrow}
\end{figure}

The first column is trained in simulation using A3C, as previously mentioned, using a wide feedforward or recurrent network. Intuitively, it makes sense to use a larger capacity network for training in simulation, to reach maximum performance. We verified this intuition by comparing wide and narrow network architectures, and found that the narrow network had slower learning and worse performance (see Figure \ref{fig:widevsnarrow}). We also see that the LSTM model out-performs the feedforward model by an average of 3 points per episode. Even on this relatively simple task, full performance is only achieved after substantial interaction with the environment, on the order of 50 million steps - a number which is infeasible with a real robot. 

The simulation training, compared with the real robot, is accelerated because of fast rendering, multithreaded learning algorithms, and the ability to continuously train without human involvement. We calculate that learning this task, which trains to convergence in 24 hours using a CPU compute cluster, would take 53 days on the real robot even with continuous training for 24 hours a day. Moreover, multiple experiments in parallel were used to explore hyperparameters in simulation; this sort of search would multiply the hypothetical real robot training time. 

In simulation, we explore learning rates and entropy costs, which are sampled uniformly at random on a log scale. Learning rates are sampled between 5e-5 and 5e-3 and entropy costs between 1e-5 and 1e-2. The configuration with the best final performance from a grid of 30 is chosen as first column.
For real Jaco experiments, both learning rates and entropy costs were optimized separately using a simulated transfer experiment with a single-threaded agent (A2C).

\subsection{Transfer to the robot}
\label{sec:transfer}

\begin{figure}
\includegraphics[width=.68\textwidth,trim={.2in, 0, .5in, 0},clip]{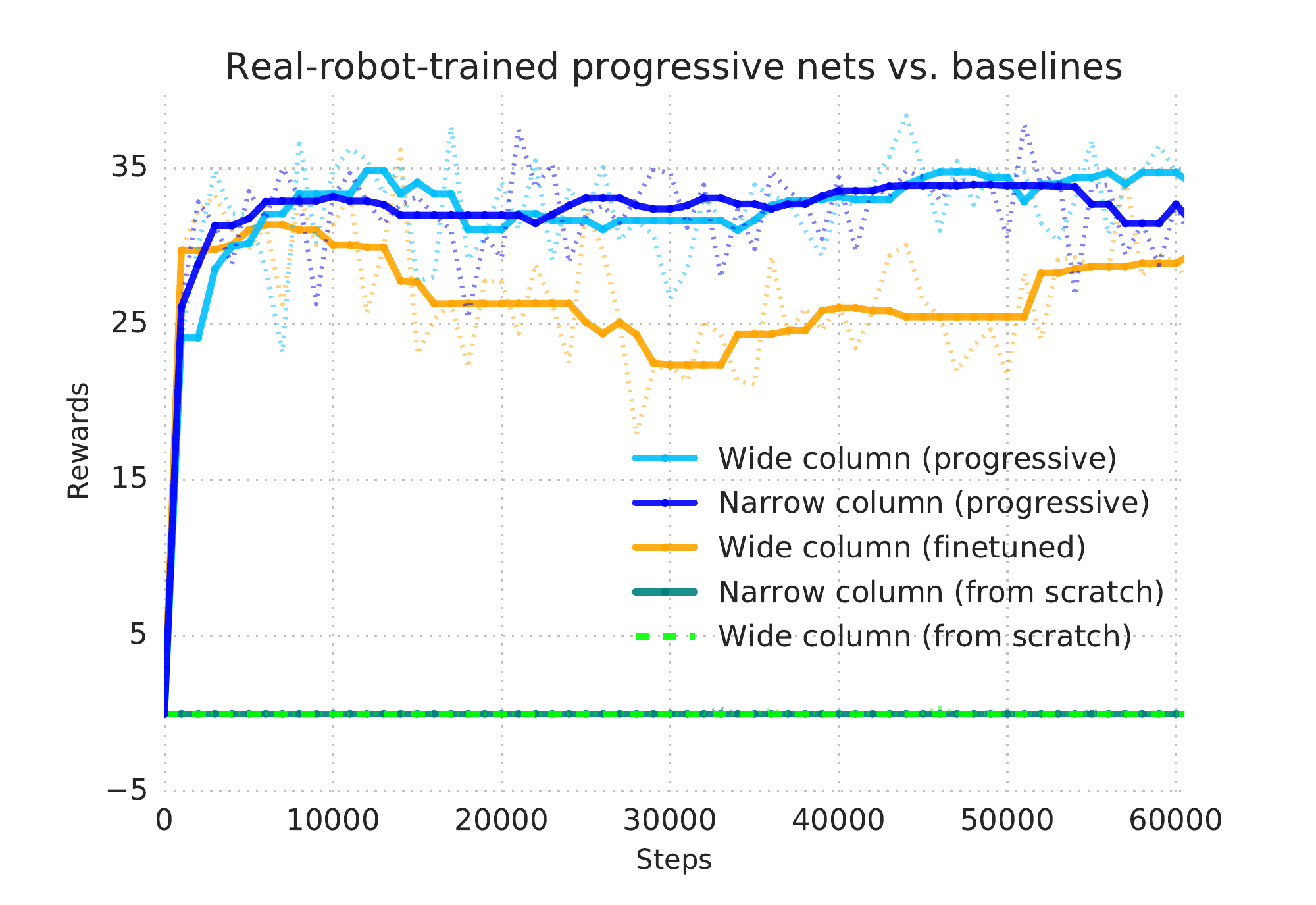}
\caption{Real robot training: We compare progressive, finetuning, and `from scratch' learning curves. All experiments use a recurrent architecture, trained on the robot, from RGB inputs. We compare wide and narrow columns for both the progressive experiments and the randomly initialized baseline. For all results, a median-filtered solid curve is shown overlaid on the raw rewards (dotted line). The `from scratch' baseline was a randomly initialized narrow or wide column, both of which fail to get any reward during training. }
\label{fig:learning_curve_col2}
\end{figure}

To train on the real Jaco, a flat target is manually repositioned within a 40cm by 30cm area on every third episode. Rewards are given automatically by tracking the colored target and giving reward based on the position of the Jaco gripper with respect to it. We train a baseline from scratch, a finetuned first column, and a progressive second column. Each experiment is run for approximately 60000 steps (about four hours). The baseline is trained by randomly initializing a narrow network and then training. We also try a randomly initialized wide network. As seen in Figure \ref{fig:learning_curve_col2} (green curve), the randomly initialized column fails to learn and the agent gets zero reward throughout training. 
The progressive second column gets to 34 points, while the experiment with finetuning, which starts with the simulation-trained column and continues training on the robot, does not reach the same score as the progressive network. 

\begin{figure}
\includegraphics[width=.24\textwidth,trim={.13in, .15in, .13in, .15in},clip]{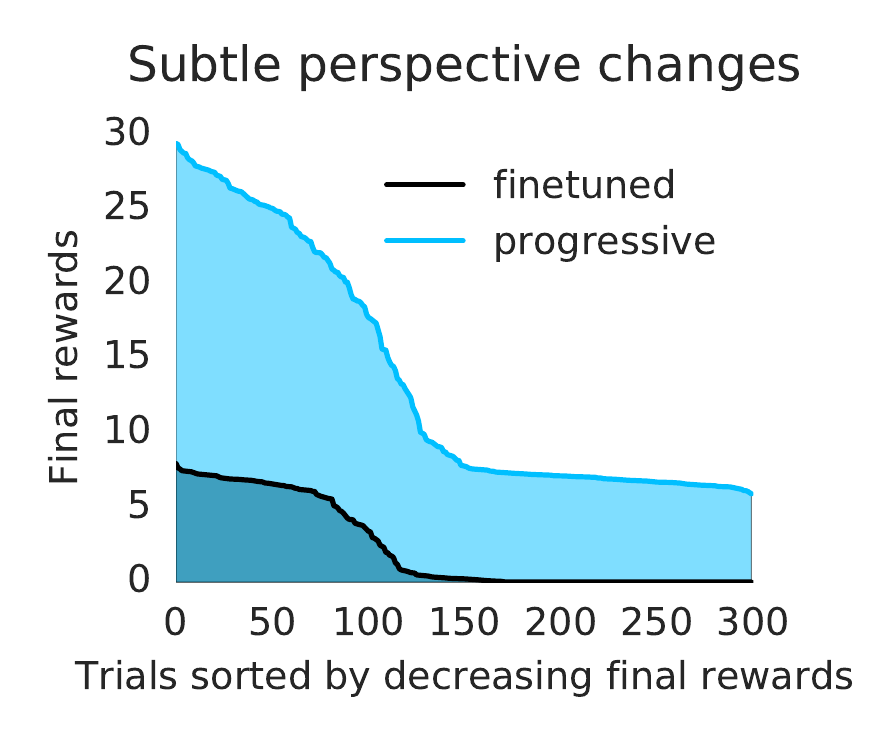}
\includegraphics[width=.24\textwidth,trim={.13in, .15in, .13in, .15in},clip]{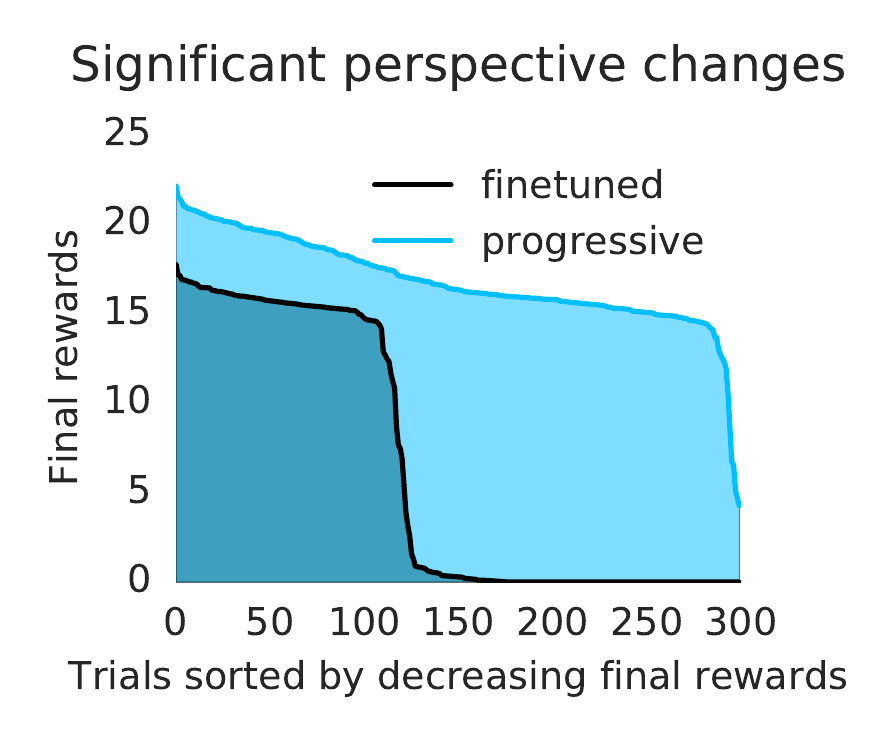}
\includegraphics[width=.24\textwidth,trim={.13in, .15in, .13in, .15in},clip]{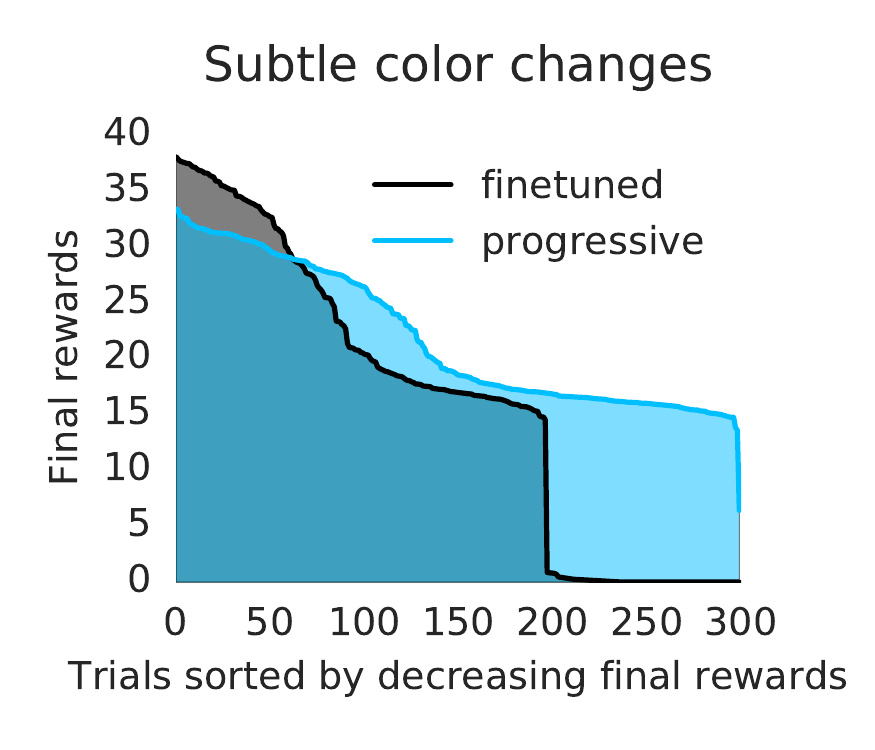}
\includegraphics[width=.24\textwidth,trim={.13in, .15in, .13in, .15in},clip]{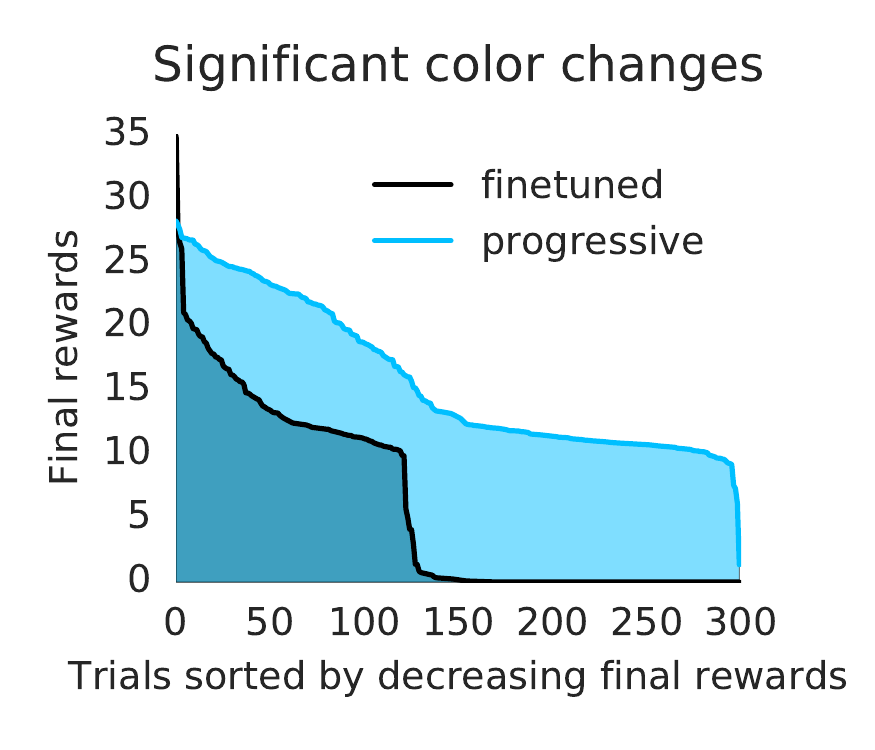} 
\caption{To analyse the relative stability and performance of finetuning vs. progressive approaches, we add color or perspective changes to the environment in simulation and then train 300 networks with different random seeds, learning rates, and entropy costs. The progressive networks have significantly higher performance and less sensitivity to hyperparameter selection for all four experiments. 
}
\label{fig:sensitivity}

\end{figure}

\textbf{Finetuning vs. progressive approaches.} The progressive approach is clearly well-suited for continual learning scenarios, where it is important to mitigate forgetting of previous tasks while supporting transfer to new tasks, but the advantage is less intuitive for curricula of tasks where the focus is on maximising transfer learning. To assess this empirically, we start with a simulator-trained first column, as described above, and then either finetune that column or add a narrow progressive column and retrain for the reacher task under a variety of conditions, including small or large color changes and small or large perspective changes. For each of these environment perturbations, we train 300 times with different seeds, learning rates, and entropy costs, which are the most sensitive hyperparameters. As shown in Figure \ref{fig:sensitivity}, we find that progressive networks are more stable and reach higher final performance than finetuning. 

\subsection{Transfer to a dynamic robot task with proprioception}

\begin{figure}
  \centering
    \includegraphics[width=.36\textwidth]{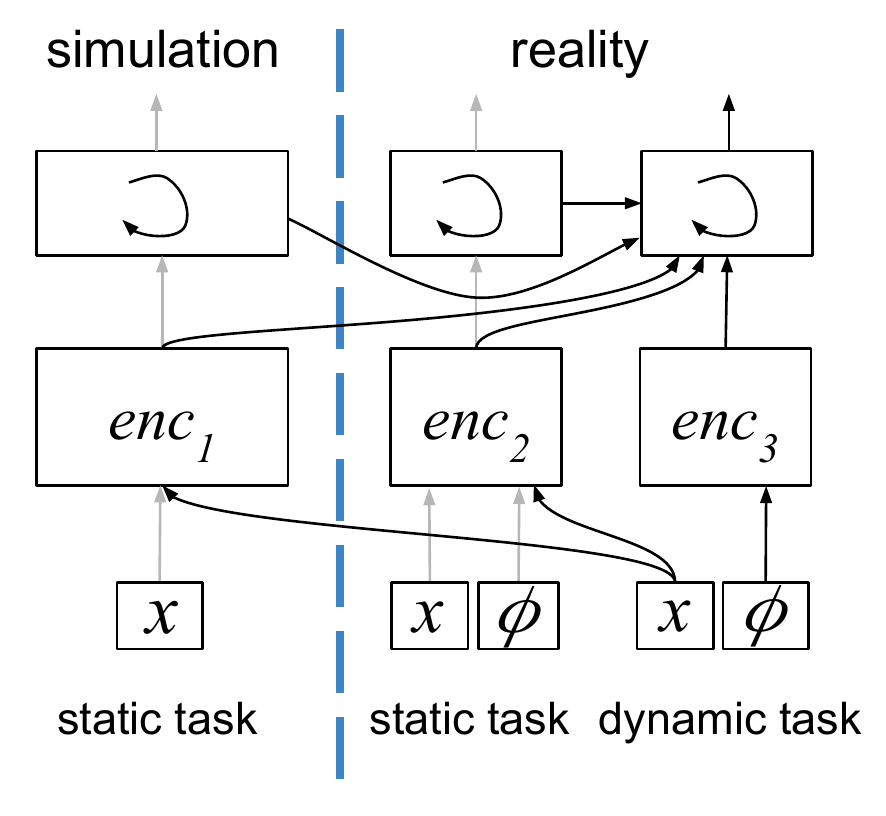}
    \includegraphics[width=.53\textwidth,trim={.2in, 0, .4in, 0},clip]{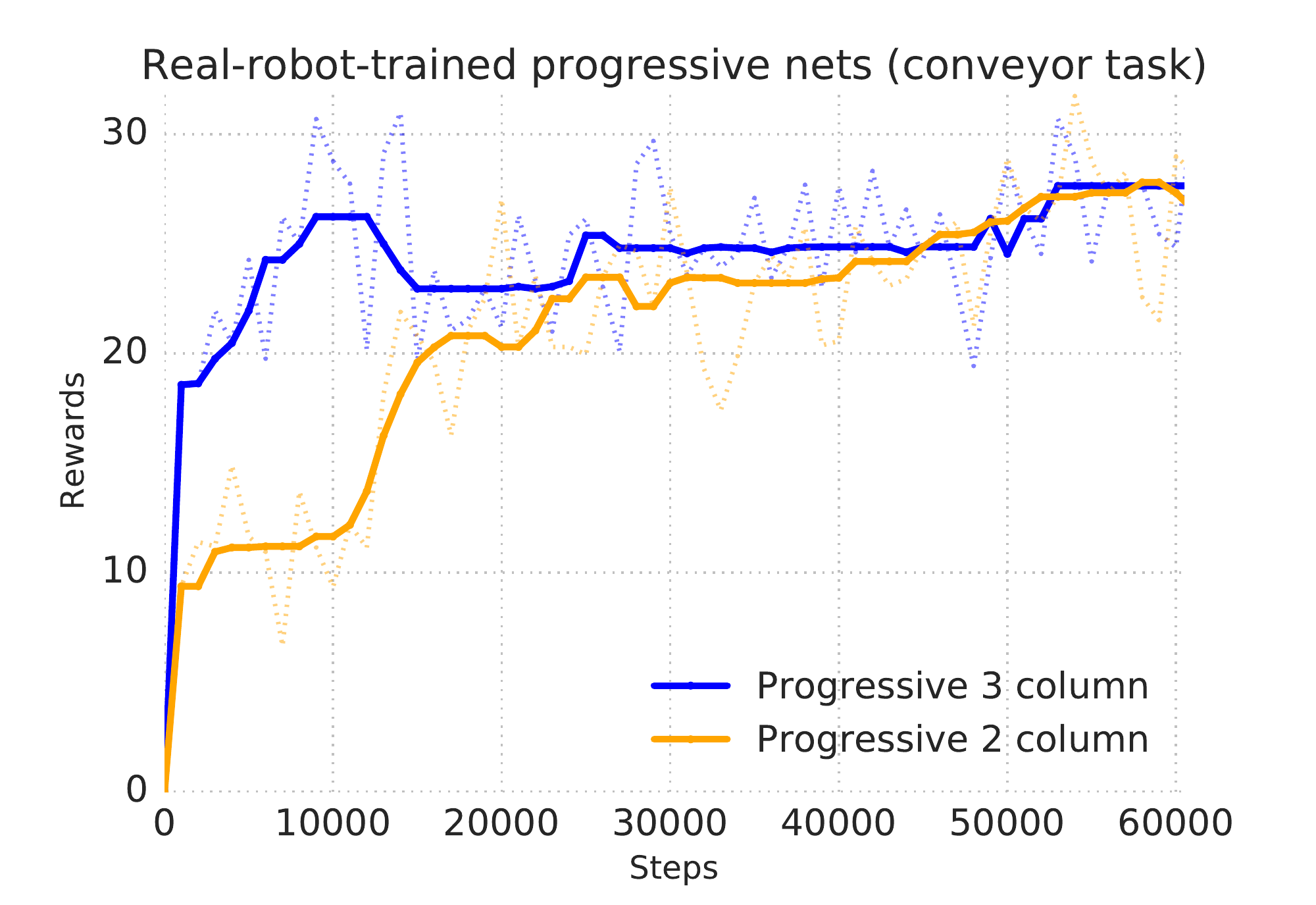}
    \caption{Real robot training results are shown for the dynamic `conveyor' task. A three-column architecture is depicted (left), in which vision ($x$) is used to train column one, vision and proprioception ($\phi$) are used in column two, and only proprioception is used to train column three. Encoder 1 is a convolutional net, encoder 2 is a convolutional net with proprioceptive features added before the LSTM, and encoder 3 is an MLP. The learning curves (right) show the results of training on a conveyor (dynamic target) task. If the conveyor task is learned as the third column, rather than the second, then the learning is significantly faster.}
    \label{fig:proprio_columns}
\end{figure}

Unlike the finetuning paradigm, which is unable to accommodate changing network morphology or new input modalities, progressive nets offer a flexibility that is advantageous for transferring to new data sources while still leveraging previous knowledge. To demonstrate this, we train a second column on the reacher task but add proprioceptive features as an additional input, alongside the RGB images. The proprioceptive features are joint angles and velocities for each of the 9 joints of the arm and fingers, 18 in total, input to a MLP (a single linear layer plus ReLU) and joined with the outputs of the convolutional stack. Then, a third progressive column is added that only learns from the proprioceptive features, while the visual input is forwarded through the previous columns and the features are used via the lateral connections. A diagram of this architecture is shown in Figure \ref{fig:proprio_columns} (left). 

To evaluate this architecture, we train on a dynamic target task. By employing a small motorized pulley, the red target is smoothly translated across the table with random reversals in the motion, creating a tracking task that requires a different control policy while maintaining a similar visual presentation. Other aspects of the task, including rewards and episode lengths, were kept the same. If the second column is trained on this \emph{conveyor} task, the learning is relatively slow, and full performance is reached after 50000 steps (about 4 hours). If the second column is instead trained on the static reacher task, and the third column is then trained on the conveyor task, we observe immediate transfer, and full performance is reached almost immediately (Figure \ref{fig:proprio_columns}, right). This demonstrates both the utility of progressive nets for curriculum tasks, as well as the capability of the architecture to immediately reuse previously learnt features. 

\section{Discussion}

Transfer learning, the ability to accumulate and transfer knowledge to new domains, is a core characteristic of intelligent beings. \textit{Progressive neural networks} offer a framework that can be used for continual learning of many tasks and which facilitates transfer learning, even across the divide which separates simulation and robot. 
We took full advantage of the flexibility and computational scaling afforded by simulation and compared many hyperparameters and architectures for a random start, random target control task with visual input, then successfully transferred the skill to an agent training on the real robot.

In order to fulfill the potential of deep reinforcement learning applied in real-world robotic domains, learning needs to become many times more efficient. One route to achieving this is via transfer learning from simulation-trained agents. We have described an initial set of experiments that prove that progressive nets can be used to achieve reliable, fast transfer for pixel-to-action RL policies.

\newpage
\footnotesize
\setlength{\bibsep}{5pt}
\bibliographystyle{plain}
\bibliography{progressive}

\begin{thebibliography}{26}
\providecommand{\natexlab}[1]{#1}
\providecommand{\url}[1]{\texttt{#1}}
\expandafter\ifx\csname urlstyle\endcsname\relax
  \providecommand{\doi}[1]{doi: #1}\else
  \providecommand{\doi}{doi: \begingroup \urlstyle{rm}\Url}\fi

\bibitem[Levine and Abbeel(2014)]{Levine2014Guided}
S.~Levine and P.~Abbeel.
\newblock Learning neural network policies with guided policy search under
  unknown dynamics.
\newblock In Z.~Ghahramani, M.~Welling, C.~Cortes, N.~D. Lawrence, and K.~Q.
  Weinberger, editors, \emph{Advances in Neural Information Processing Systems
  27}, pages 1071--1079. Curran Associates, Inc., 2014.
\newblock URL
  \url{http://papers.nips.cc/paper/5444-learning-neural-network-policies-with-guided-policy-search-under-unknown-dynamics.pdf}.

\bibitem[Schulman et~al.(2015)Schulman, Levine, Moritz, Jordan, and
  Abbeel]{Schulmanetal_ICML2015}
J.~Schulman, S.~Levine, P.~Moritz, M.~I. Jordan, and P.~Abbeel.
\newblock Trust region policy optimization.
\newblock In \emph{Proceedings of the 32nd International Conference on Machine
  Learning (ICML)}, 2015.

\bibitem[Heess et~al.(2015)Heess, Wayne, Silver, Lillicrap, Erez, and
  Tassa]{Heess2015Stochastic}
N.~Heess, G.~Wayne, D.~Silver, T.~P. Lillicrap, T.~Erez, and Y.~Tassa.
\newblock Learning continuous control policies by stochastic value gradients.
\newblock In \emph{Advances in Neural Information Processing Systems 28: Annual
  Conference on Neural Information Processing Systems 2015, December 7-12,
  2015, Montreal, Quebec, Canada}, pages 2944--2952, 2015.
\newblock URL
  \url{http://papers.nips.cc/paper/5796-learning-continuous-control-policies-by-stochastic-value-gradients}.

\bibitem[Lillicrap et~al.(2016)Lillicrap, Hunt, Pritzel, Heess, Erez, Tassa,
  Silver, and Wierstra]{Lillicrap2016Continuous}
T.~P. Lillicrap, J.~J. Hunt, A.~Pritzel, N.~Heess, T.~Erez, Y.~Tassa,
  D.~Silver, and D.~Wierstra.
\newblock Continuous control with deep reinforcement learning.
\newblock \emph{Proceedings of the International Conference on Learning
  Representations (ICLR)}, 2016.
\newblock URL \url{http://arxiv.org/abs/1509.02971}.

\bibitem[Schulman et~al.(2016)Schulman, Moritz, Levine, Jordan, and
  Abbeel]{Schulmanetal_ICLR2016}
J.~Schulman, P.~Moritz, S.~Levine, M.~Jordan, and P.~Abbeel.
\newblock High-dimensional continuous control using generalized advantage
  estimation.
\newblock In \emph{Proceedings of the International Conference on Learning
  Representations (ICLR)}, 2016.

\bibitem[Mnih et~al.(2016)Mnih, Badia, Mirza, Graves, Lillicrap, Harley,
  Silver, and Kavukcuoglu]{mnih2016a3c}
V.~Mnih, A.~P. Badia, M.~Mirza, A.~Graves, T.~P. Lillicrap, T.~Harley,
  D.~Silver, and K.~Kavukcuoglu.
\newblock Asynchronous methods for deep reinforcement learning.
\newblock In \emph{Int'l Conf. on Machine Learning (ICML)}, 2016.

\bibitem[Gu et~al.(2016)Gu, Lillicrap, Sutskever, and Levine]{Gu2016NAF}
S.~Gu, T.~P. Lillicrap, I.~Sutskever, and S.~Levine.
\newblock Continuous deep q-learning with model-based acceleration.
\newblock In \emph{ICML 2016}, 2016.

\bibitem[Rusu et~al.(2016)Rusu, Rabinowitz, Desjardins, Soyer, Kirkpatrick,
  Kavukcuoglu, Pascanu, and Hadsell]{rusu2016progressive}
A.~Rusu, N.~Rabinowitz, G.~Desjardins, H.~Soyer, J.~Kirkpatrick,
  K.~Kavukcuoglu, R.~Pascanu, and R.~Hadsell.
\newblock Progressive neural networks.
\newblock \emph{arXiv preprint arXiv:1606.04671}, 2016.

\bibitem[Peng et~al.(2015)Peng, Sun, Ali, and Saenko]{PengSAS15}
X.~Peng, B.~Sun, K.~Ali, and K.~Saenko.
\newblock Learning deep object detectors from 3d models.
\newblock In \emph{2015 {IEEE} International Conference on Computer Vision,
  {ICCV} 2015, Santiago, Chile, December 7-13, 2015}, pages 1278--1286, 2015.

\bibitem[Su et~al.(2015)Su, Qi, Li, and Guibas]{SuQLG15}
H.~Su, C.~R. Qi, Y.~Li, and L.~J. Guibas.
\newblock Render for {CNN:} viewpoint estimation in images using cnns trained
  with rendered 3d model views.
\newblock In \emph{2015 {IEEE} International Conference on Computer Vision,
  {ICCV} 2015, Santiago, Chile, December 7-13, 2015}, pages 2686--2694, 2015.

\bibitem[Long et~al.(2015)Long, Cao, Wang, and Jordan]{LongC0J15}
M.~Long, Y.~Cao, J.~Wang, and M.~I. Jordan.
\newblock Learning transferable features with deep adaptation networks.
\newblock In \emph{Proceedings of the 32nd International Conference on Machine
  Learning, {ICML} 2015, Lille, France, 6-11 July 2015}, pages 97--105, 2015.

\bibitem[Tzeng et~al.(2015)Tzeng, Hoffman, Darrell, and Saenko]{TzengHDS15}
E.~Tzeng, J.~Hoffman, T.~Darrell, and K.~Saenko.
\newblock Simultaneous deep transfer across domains and tasks.
\newblock In \emph{2015 {IEEE} International Conference on Computer Vision,
  {ICCV} 2015, Santiago, Chile, December 7-13, 2015}, pages 4068--4076, 2015.

\bibitem[Tzeng et~al.(2014)Tzeng, Hoffman, Zhang, Saenko, and
  Darrell]{TzengHZSD14}
E.~Tzeng, J.~Hoffman, N.~Zhang, K.~Saenko, and T.~Darrell.
\newblock Deep domain confusion: Maximizing for domain invariance.
\newblock \emph{CoRR}, abs/1412.3474, 2014.
\newblock URL \url{http://arxiv.org/abs/1412.3474}.

\bibitem[Tzeng et~al.(2015)Tzeng, Devin, Hoffman, Finn, Peng, Levine, Saenko,
  and Darrell]{TzengDHFPLSD15}
E.~Tzeng, C.~Devin, J.~Hoffman, C.~Finn, X.~Peng, S.~Levine, K.~Saenko, and
  T.~Darrell.
\newblock Towards adapting deep visuomotor representations from simulated to
  real environments.
\newblock \emph{CoRR}, abs/1511.07111, 2015.
\newblock URL \url{http://arxiv.org/abs/1511.07111}.

\bibitem[Ganin et~al.(2016)Ganin, Ustinova, Ajakan, Germain, Larochelle,
  Laviolette, Marchand, and Lempitsky]{ganin2015domain}
Y.~Ganin, E.~Ustinova, H.~Ajakan, P.~Germain, H.~Larochelle, F.~Laviolette,
  M.~Marchand, and V.~Lempitsky.
\newblock Domain-adversarial training of neural networks.
\newblock \emph{Journal of Machine Learning Research}, 17\penalty0
  (59):\penalty0 1--35, 2016.

\bibitem[Ajakan et~al.(2014)Ajakan, Germain, Larochelle, Laviolette, and
  Marchand]{ajakanDomain2016}
H.~Ajakan, P.~Germain, H.~Larochelle, F.~Laviolette, and M.~Marchand.
\newblock Domain-adversarial neural networks.
\newblock \emph{CoRR}, abs/1412.4446, 2014.
\newblock URL \url{http://arxiv.org/abs/1412.4446}.

\bibitem[Bousmalis et~al.(2016)Bousmalis, Trigeorgis, Silberman, Krishnan, and
  Erhan]{BousmalisTSKE16}
K.~Bousmalis, G.~Trigeorgis, N.~Silberman, D.~Krishnan, and D.~Erhan.
\newblock Domain separation networks.
\newblock In \emph{Advances in Neural Information Processing Systems}, pages
  343--351, 2016.

\bibitem[Barrett et~al.(2010)Barrett, Taylor, and Stone]{AAMASWS10-barrett}
S.~Barrett, M.~E. Taylor, and P.~Stone.
\newblock Transfer learning for reinforcement learning on a physical robot.
\newblock In \emph{Ninth International Conference on Autonomous Agents and
  Multiagent Systems - Adaptive Learning Agents Workshop (AAMAS - ALA)}, 2010.

\bibitem[{James} and {Johns}(2016)]{2016JamesJohns}
S.~{James} and E.~{Johns}.
\newblock {3D Simulation for Robot Arm Control with Deep Q-Learning}.
\newblock \emph{ArXiv e-prints}, 2016.

\bibitem[Zhu et~al.(2017)Zhu, Mottaghi, Kolve, Lim, Gupta, Fei-Fei, and
  Farhadi]{ZhuMKLGFF16}
Y.~Zhu, R.~Mottaghi, E.~Kolve, J.~J. Lim, A.~Gupta, L.~Fei-Fei, and A.~Farhadi.
\newblock Target-driven visual navigation in indoor scenes using deep
  reinforcement learning.
\newblock In \emph{Robotics and Automation (ICRA), 2017 IEEE International
  Conference on}, pages 3357--3364. IEEE, 2017.

\bibitem[Levine et~al.(2016)Levine, Finn, Darrell, and Abbeel]{LevineFDA15}
S.~Levine, C.~Finn, T.~Darrell, and P.~Abbeel.
\newblock End-to-end training of deep visuomotor policies.
\newblock \emph{Journal of Machine Learning Research}, 17\penalty0
  (39):\penalty0 1--40, 2016.

\bibitem[Levine et~al.(2015)Levine, Wagener, and Abbeel]{LevineWA15}
S.~Levine, N.~Wagener, and P.~Abbeel.
\newblock Learning contact-rich manipulation skills with guided policy search.
\newblock In \emph{{IEEE} International Conference on Robotics and Automation,
  {ICRA} 2015, Seattle, WA, USA, 26-30 May, 2015}, pages 156--163, 2015.

\bibitem[Pinto and Gupta(2016)]{PintoG16}
L.~Pinto and A.~Gupta.
\newblock Supersizing self-supervision: Learning to grasp from 50k tries and
  700 robot hours.
\newblock In \emph{ICRA 2016}, 2016.

\bibitem[Levine et~al.(2016)Levine, Pastor, Krizhevsky, Ibarz, and
  Quillen]{levine2016Large}
S.~Levine, P.~Pastor, A.~Krizhevsky, J.~Ibarz, and D.~Quillen.
\newblock Learning hand-eye coordination for robotic grasping with deep
  learning and large-scale data collection.
\newblock \emph{The International Journal of Robotics Research}, page
  0278364917710318, 2016.

\bibitem[Mnih et~al.(2015)Mnih, Kavukcuoglu, Silver, Rusu, Veness, Bellemare,
  Graves, Riedmiller, Fidjeland, Ostrovski, Petersen, Beattie, Sadik,
  Antonoglou, King, Kumaran, Wierstra, Legg, and Hassabis]{mnih-dqn-2015}
V.~Mnih, K.~Kavukcuoglu, D.~Silver, A.~Rusu, J.~Veness, M.~Bellemare,
  A.~Graves, M.~Riedmiller, A.~Fidjeland, G.~Ostrovski, S.~Petersen,
  C.~Beattie, A.~Sadik, I.~Antonoglou, H.~King, D.~Kumaran, D.~Wierstra,
  S.~Legg, and D.~Hassabis.
\newblock Human-level control through deep reinforcement learning.
\newblock \emph{Nature}, 518\penalty0 (7540):\penalty0 529--533, 2015.

\bibitem[Todorov et~al.(2012)Todorov, Erez, and Tassa]{Todorov2012}
E.~Todorov, T.~Erez, and Y.~Tassa.
\newblock Mujoco: A physics engine for model-based control.
\newblock In \emph{International Conference on Intelligent Robots and Systems
  {IROS}}, 2012.

\end{thebibliography}

\newpage

\appendix
\pagenumbering{arabic}

\end{document}